%% file: PaperForReview.tex
\documentclass[10pt,twocolumn,letterpaper]{article}
\pdfoutput=1    % For arXiv issues

\usepackage[pagenumbers]{cvpr} % To force page numbers, e.g. for an arXiv version

\makeatletter
\@namedef{ver@everyshi.sty}{}
\makeatother

\usepackage[accsupp]{axessibility}  % Improves PDF readability for those with disabilities.

\usepackage{graphicx}
\usepackage{amsmath}
\usepackage{amssymb}
\usepackage{booktabs}

\usepackage{svg}
\usepackage{cite}
\usepackage{amsmath,amssymb,amsfonts}
\usepackage{algorithmic}
\usepackage{textcomp}
\usepackage{xcolor}
\usepackage{pifont} %xmark
\usepackage{float}
\usepackage{siunitx}
\usepackage[nolist,nohyperlinks]{acronym}
\usepackage{url}

\usepackage[pagebackref,breaklinks,colorlinks]{hyperref}

\usepackage[capitalize]{cleveref}
\crefname{section}{Sec.}{Secs.}
\Crefname{section}{Section}{Sections}
\Crefname{table}{Table}{Tables}
\crefname{table}{Tab.}{Tabs.}

\def\cvprPaperID{22} % *** Enter the CVPR Paper ID here
\def\confName{CVPR}
\def\confYear{2022}

\newcommand{\cmark}{\ding{51}}%
\newcommand{\xmark}{\ding{55}}%

\acrodef{SemSeg}[SemSeg]{semantic segmentation}
\acrodef{AD}[AD]{autonomous driving}
\acrodef{LaF}[LaF]{Lost-and-Found}
\acrodef{JS}[JS]{Jensen-Shannon}
\acrodef{PSNR}[PSNR]{signal-to-noise ratio}
\acrodef{DM}[DM]{domain mismatch}
\acrodef{MSE}[MSE]{mean-squared error}
\acrodef{EDS}[EDS]{euclidean distance sum}
\acrodef{BEV}[BEV]{bird's eye view}
\acrodef{NLOS}[NLOS]{Non-Line-of-Sight}
\acrodef{ML}[ML]{Machine Learning}
\acrodef{DL}[DL]{Deep Learning}
\acrodef{MSG}[MSG]{multi-scale grouping}
\acrodef{DUT}[DUT]{detection under test}
\acrodef{OGM}[OGM]{occupancy grid map}
\acrodef{SSIM}[SSIM]{structural similarity index measure}
\acrodef{EMD}[EMD]{earth-mover's distance}
\acrodef{EMDEV}[EMDEV]{earth-mover's deviation}
\acrodef{ID}[ID]{in-distribution}
\acrodef{OOD}[OOD]{out-of-distribution}
\acrodef{CS}[CS]{Cityscapes}
\acrodef{FS}[FS]{Fishyscapes}
\acrodef{SMIYC}[SMIYC]{Segment Me If You Can}

\begin{document}

\title{Anomaly Detection in Autonomous Driving: A Survey}

\author{Daniel Bogdoll$^{1,2}$\textsuperscript{\textasteriskcentered}, Maximilian Nitsche$^{1,2}$\textsuperscript{\textasteriskcentered}, J. Marius Zöllner$^{1,2}$\\
$^{1}$FZI Research Center for Information Technology, Karlsruhe, Germany\\
$^{2}$KIT Karlsruhe Institute of Technology, Karlsruhe, Germany\\
{\tt\small \{bogdoll, nitsche, zoellner\}@fzi.de}
}

\maketitle

\begingroup\renewcommand\thefootnote{\textasteriskcentered}
\footnotetext{These authors contributed equally}
\endgroup

%%%%%%%%% ABSTRACT
\begin{abstract}
Nowadays, there are outstanding strides towards a future with autonomous vehicles on our roads. While the perception of autonomous vehicles performs well under closed-set conditions, they still struggle to handle the unexpected. This survey provides an extensive overview of anomaly detection techniques based on camera, lidar, radar, multimodal and abstract object level data. We provide a systematization including detection approach, corner case level, ability for an online application, and further attributes. We outline the state-of-the-art and point out current research gaps.
\end{abstract}

\input{sections/0_introduction}

\input{sections/1_camera}
\input{sections/2_lidar}
\input{sections/3_radar}

\input{sections/4_multimodal}
\input{sections/5_object}
\input{sections/6_conclusion}

\section*{Acknowledgment}
This work results from the project KI Data Tooling (19A20001J), funded by the German Federal Ministry for Economic Affairs and Climate Action (BMWK).

%%%%%%%%% REFERENCES
{\small
\bibliographystyle{ieee_fullname}
\bibliography{references}
}

\end{document}

%% file: sections/0_introduction.tex
\section{Introduction}

Anomalies, also called corner cases, occur everyday on the street, which is why autonomous vehicles need to cope with them. This “long tail of rare events”~\cite{jainAutonomy2021} is seen by many as the core obstacle towards large scale deployments of autonomous vehicles~\cite{anguelovMIT2019, schneiderCMU2019, karpathyTesla2019}. While there are exciting advances in handling the rare and unknown~\cite{wang2022freesolo, wang2020frustratingly, josephOpenWorldObject2021}, it remains crucial to detect anomalies, which is still challenging~\cite{LiCODA2022}. In \ac{AD}, there are many levels of corner cases and multiple sensor modalities, including camera, lidar, and radar. While an extensive survey regarding camera-based approaches~\cite{breitensteinCornerCasesVisual2021} exists, there is little to no research regarding other sensors or corner cases on higher levels of abstraction, including surveys. Here, we provide an overview of anomaly detection methods in the domain of \ac{AD} for different sensor modalities, including methods not explicitly developed for \ac{AD}, but which we deem applicable. 

\begin{figure}[t!]
    \centering
    \resizebox{\linewidth}{!}{
    \includegraphics{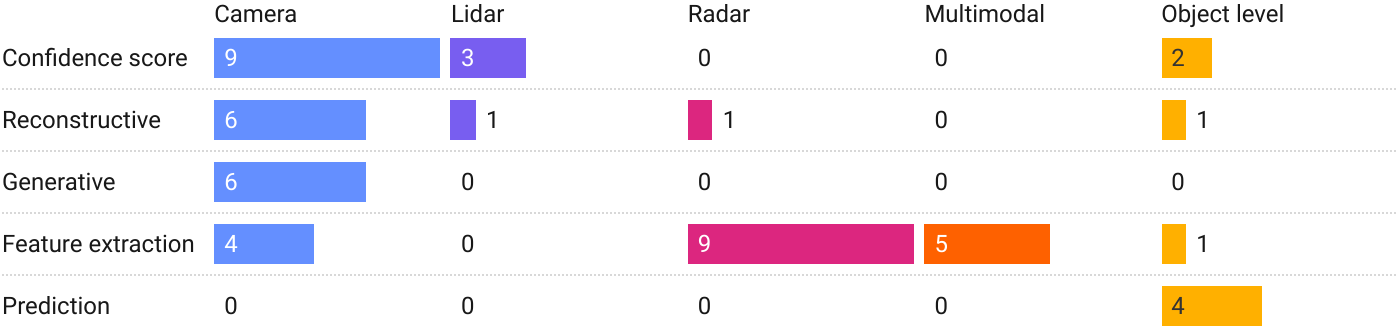}}
    \caption{Overview of anomaly detection approaches based on camera, lidar, radar, multimodal, and abstract object level data.}
    \label{fig:bar_chart_approaches}
\end{figure}

%Systemization of anomaly detection techniques
We characterize the anomaly detection techniques in Tables~\ref{tab:cameraOverview}-\ref{tab:abstraction_object_overview} across the modalities camera, lidar, radar, multimodal, and abstract object level. They are further characterized by their general detection approach, type of corner case, evaluation dataset or simulation, as well as regarding their possible online application. We classify the detection approaches following Breitenstein~et~al. in five concepts: “reconstruction, prediction, generative, confidence scores, and feature extraction”~\cite{breitensteinCornerCasesVisual2021}. Confidence score techniques are often derived by post-processing without interfering with the training of a neural network and subdivided into Bayesian approaches, learned scores, and scores obtained by post-processing. Reconstructive approaches try to reconstruct normality and consider any kind of deviation from it as anomalous. Generative approaches are closely related to the former reconstructive approaches, but also take into account the discriminator's decision or the distance to the training data. Feature extraction can be based on handcrafted or learned features to determine a class label or compare modalities on various feature levels. Prediction based techniques predict the next frame(s) expected under normality. An overview can be found in Figure~\ref{fig:bar_chart_approaches}. 
%Inclusion and exclusion criteria
% exclude approaches which analyze the sample in contrast to the training distribution

We follow Breitenstein~et~al.~\cite{breitensteinSystematizationCornerCases2020} for the systematization of corner cases with the levels pixel, domain, object, scene and scenario, each being harder to detect. Heidecker~et~al.~\cite{heideckerApplicationDrivenConceptualizationCorner2021} extended these camera-based levels to incorporate lidar and radar sensors. Similar to their work, we use the terms “anomaly” and “corner case” interchangeably. In this survey, we focus on natural, external corner cases. Thus, we exclude anomalies on the sensor layer \cite{heideckerApplicationDrivenConceptualizationCorner2021}; anomalies on the pixel level; and anomalies due to adversarial attacks. %As some anomaly datasets \cite{blumFishyscapesBenchmarkMeasuring2021, hendrycksScalingOutofDistributionDetection2022} tackle obstacles on the road, we also consider road obstacles as possible corner cases, considering them as contextual anomalies. Specifically, some road objects might have been in the training set but not located on the road.

We list all datasets or simulation environments used and label techniques as online capable if they or similar approaches of equal computational complexity (or higher, denoted by **) are reported as such or name a frame rate above 10 FPS. Methods marked with * are not providing inference performance measurements and are thus labeled as offline.

%% file: sections/1_camera.tex
\section{Anomaly Detection on Camera Data}

% Ideen für Eingrenzung:
% - CVPR, ICCV, ECCV, NeurIPS, IROS, ICRA, ACCV, BMVC, ICPR (https://personalinterests.lipingyang.org/wp-content/uploads/2019/03/How-to-get-your-CVPR-paper-rejected.pptx.pdf)
% - IV, ITSC (https://2021.ieee-iv.org/, https://2021.ieee-itsc.org/)
% - mit h5 Index könnte man ITSC rausargumentieren
% - ab 2019
% - Benchmark Performance (e.g. https://fishyscapes.com/results)
% - future frame predictions? Video based detection?
% - Fahrerverhalten
% - Recognizing an unknown as an unknown requires strong generalization
% - Filtered by Lost & Found dataset AP benchmark
% - above 10\% AP
% - Focus on Road obstacles = Scene Level

Autonomous vehicles are often equipped with different camera systems, like stereo, mono, and fisheye cameras, to ensure a rich perception of the environment. Thus, anomaly detection in camera data holds great potential for more robust visual perception. For this section, we introduce two more criteria following the \ac{FS} benchmark~\cite{fishyscapesResultsFishyscapesBenchmark2022}: auxiliary data and retraining. The former indicates whether an approach requires anomalous data during training. Retraining, however, specifies whether methods cannot use pretrained models, but require a special loss or the retraining, which might decrease the performance~\cite{fishyscapesResultsFishyscapesBenchmark2022}. All camera-based methods can be found in Table~\ref{tab:cameraOverview}.

%TODO remove references of datasets and pool them in footnote
\begin{table*}[ht]
\caption{Overview of anomaly detection techniques on camera data.}
\label{tab:cameraOverview}
\resizebox{\textwidth}{!}{
\begin{tabular}{@{}lccllcclll@{}}
\toprule
Author(s) & Year & Ref & Technique &  Approach & Aux Data & Retraining & Corner Case Level  & Dataset / Simulation & Online\\
\midrule
Du~et~al.~& 2022 &\cite{duVOSLearningWhat2022} & VOS & Confidence --- Learned &\xmark & \cmark& Object  --- Single-Point & PASCAL-VOC\cite{everinghamPascalVisualObject2010}, BDD100K\cite{yuBDD100KDiverseDriving2020}& \xmark* \\
Jung~et~al.~&2021&\cite{jungStandardizedMaxLogits2021a}&Standardized Max Logits&Confidence --- Learned &\xmark&\xmark&Scene  --- Contextual &\ac{FS} \ac{LaF}\cite{blumFishyscapesBenchmarkSafe2019}, RA\cite{lisDetectingUnexpectedImage2019}&\cmark (13.3 FPS)\\
Heidecker~et~al.~&2021&\cite{heideckerCornerCaseDetection2021}&MC Dropout&Confidence --- Bayesian & \xmark&\xmark&Object  --- Single Point&A2D2\cite{AudiAGDrivingDataset2019}&\xmark\\
Chan~et~al.~&2021&\cite{chanEntropyMaximizationMeta2021a}&Entropy Maximization&Confidence --- Learned & \cmark&\cmark&Object  --- Single-Point&\ac{LaF}\cite{pinggeraLostFoundDetecting2016a}, \ac{CS}\cite{cordtsCityscapesDatasetSemantic2016}, \ac{FS}\cite{blumFishyscapesBenchmarkSafe2019} & \cmark\\
Breitenstein~et~al.~&2021& \cite{breitensteinDetectionCollectiveAnomalies2021} & EMDEV & Confidence --- Post-processed & \xmark & \cmark & Scene  --- Collective & \ac{CS}\cite{cordtsCityscapesDatasetSemantic2016} as $D_{in}$, ECP\cite{braunEuroCityPersonsNovel2019} \& A2D2\cite{AudiAGDrivingDataset2019} as $D_{out}$& \cmark\\
Bevandić~et~al.~&2021&\cite{bevandicDenseOutlierDetection2021}&LDN-BIN&Confidence --- Learned&\cmark&\cmark&Object  --- Single-Point&Vistas\cite{neuholdMapillaryVistasDataset2017a} \& \ac{CS}\cite{cordtsCityscapesDatasetSemantic2016} as $D_{in}$, ImageNet\cite{dengImageNetLargescaleHierarchical2009} \& WD\cite{bevandicSimultaneousSemanticSegmentation2019a} as $D_{out}$&\xmark*\\
Malinin and Gales &2018&\cite{malininPredictiveUncertaintyEstimation2018a}&Dirichlet Prior Networks&Confidence --- Bayesian&\cmark&\cmark&Object  --- Contextual&\ac{FS} \ac{LaF}\cite{blumFishyscapesBenchmarkSafe2019}&\xmark*\\
Huang~et~al.~&2018&\cite{huangEfficientUncertaintyEstimation2018}&RTA&Confidence --- Bayesian &\xmark&\xmark&Object  --- Single-Point&CamVid\cite{brostowSemanticObjectClasses2009}&\cmark \\
Kendall~et~al.~&2016&\cite{kendallBayesianSegNetModel2016} & Bayesian SegNet & Confidence --- Bayesian & \xmark & \xmark & Object --- Single-Point & CamVid\cite{brostowSemanticObjectClasses2009} & \cmark (11.11 FPS)\\
Vojir~et~al.~& 2021 & \cite{vojirRoadAnomalyDetection2021a}&JSR-Net&Reconstruction&\xmark&\xmark&Scene  --- Contextual&\ac{LaF}\cite{pinggeraLostFoundDetecting2016a}, RA\cite{lisDetectingUnexpectedImage2019} \& RO\cite{lisDetectingRoadObstacles2021}, \ac{FS} \ac{LaF}\cite{blumFishyscapesBenchmarkSafe2019} &\xmark*\\
Ohgushi~et~al.~& 2021 &\cite{ohgushiRoadObstacleDetection2021} &Autoencoder + SemSeg &Reconstruction&\xmark&\xmark&Scene  --- Contextual&\ac{LaF}\cite{pinggeraLostFoundDetecting2016a}, Highway dataset & \xmark\\
Lis~et~al.~&2021&\cite{lisDetectingRoadObstacles2021}& Erasing &Reconstruction&\xmark&\cmark&Scene  --- Contextual& \ac{FS} \ac{LaF}\cite{blumFishyscapesBenchmarkSafe2019}, RO (daylight)\cite{lisDetectingRoadObstacles2021}& \xmark*\\
Di Biase~et~al.~&2021&\cite{dibiasePixelwiseAnomalyDetection2021a}& SynBoost &Reconstruction &\cmark&\xmark&Object  --- Single-Point& \ac{FS} \ac{LaF}, Static, Web (Oct. 2020)\cite{blumFishyscapesBenchmarkSafe2019}&\xmark\\
Blum~et~al.~&2021&\cite{blumFishyscapesBenchmarkMeasuring2021}&NF + Logistic regression&Reconstruction&\cmark&\xmark&Object  --- Single-Point&\ac{FS} \ac{LaF}\cite{blumFishyscapesBenchmarkSafe2019}, \ac{FS} Web \& Static\cite{blumFishyscapesBenchmarkSafe2019}&\xmark\\
Creusot and Munawar &2015& \cite{creusotRealtimeSmallObstacle2015} & Compressive RBM & Reconstruction & \xmark & \cmark & Scene  --- Contextual & Recordings \& YT Japanese highways & \cmark (10 FPS) \\
Nitsch~et~al.~&2021 &\cite{nitschOutofDistributionDetectionAutomotive2021}&GAN + Post hoc statistics&Generative&\xmark&\cmark&Object  --- Single-Point&KITTI\cite{geigerVisionMeetsRobotics2013} \& nuScenes\cite{caesarNuScenesMultimodalDataset2020} as $D_{in}$, ImageNet\cite{dengImageNetLargescaleHierarchical2009} as $D_{out}$ &\cmark\\
Grcić~et~al.~&2021&\cite{grcicDenseAnomalyDetection2021}&NFlowJS&Generative&\xmark&\cmark&Object  --- Single-Point&WD-Pascal\cite{bevandicSimultaneousSemanticSegmentation2019a}, \ac{LaF}\cite{pinggeraLostFoundDetecting2016a}, SMIYC\cite{chanSegmentMeIfYouCanBenchmarkAnomaly2021}, StreetHazards\cite{hendrycksScalingOutofDistributionDetection2022}&\cmark (18.4 FPS)\\
Xia~et~al.~&2020& \cite{xiaSynthesizeThenCompare2020} &SynthCP&Generative&\xmark&\xmark&Object  --- Single-Point&\ac{CS}\cite{cordtsCityscapesDatasetSemantic2016}, StreetHazards\cite{hendrycksScalingOutofDistributionDetection2022}&\xmark*\\
Löhdefink~et~al.~&2020&\cite{lohdefinkSelfSupervisedDomainMismatch2020}&Autoencoder DM&Generative&\xmark&\cmark&Domain  --- Domain Shift&\ac{CS}\cite{cordtsCityscapesDatasetSemantic2016}, BDD100K\cite{yuBDD100KDiverseDriving2020}, KITTI\cite{geigerVisionMeetsRobotics2013}&\cmark\\
Lis~et~al.~&2019&\cite{lisDetectingUnexpectedImage2019}&Resynthesis&Generative&\xmark&\xmark&Object  --- Single-Point&\ac{LaF}\cite{pinggeraLostFoundDetecting2016a}, RA\cite{lisDetectingUnexpectedImage2019}& \xmark*\\
Haldimann~et~al.~&2019& \cite{haldimannThisNotWhat2019}&Semantic cGAN&Generative&\xmark&\xmark&Scene  --- Single-Point&\ac{CS}\cite{cordtsCityscapesDatasetSemantic2016} as $D_{in}$, Vistas\cite{neuholdMapillaryVistasDataset2017a} as $D_{out}$ \cite{neuholdMapillaryVistasDataset2017}&\xmark*\\
Xue~et~al.~&2019&\cite{xueNovelMultilayerFramework2019}&Multi-layer Occlusion&Feature Extraction&\cmark&\cmark&Scene  --- Contextual&\ac{LaF}\cite{pinggeraLostFoundDetecting2016a}&\cmark\\
Bolte~et~al.~&2019&\cite{bolteUnsupervisedDomainAdaptation2019} &Feature MSE&Feature Extraction&\xmark&\cmark&Domain  --- Domain Shift&KITTI\cite{geigerVisionMeetsRobotics2013}, \ac{CS}\cite{cordtsCityscapesDatasetSemantic2016}, BDD100K\cite{yuBDD100KDiverseDriving2020} &\cmark\\
Zhang~et~al.~&2018&\cite{zhangDeepRoadGANbasedMetamorphic2018} & DeepRoad & Feature Extraction&\xmark&\xmark&Domain  --- Domain Shift & Udacity\cite{FinalLeaderboardUdacity2016} & \cmark \\
Bai~et~al.~&2018&\cite{baiRecognizingAnomaliesUrban2018}&SVM&Feature Extraction&\cmark&\xmark&Scene  --- Contextual \& Collective& Urban roads & \cmark\\
\bottomrule
\end{tabular}}
\end{table*}

%Camera baseline
\textbf{Confidence score.} Approaches on the basis of confidence scores constitute a baseline for the detection of anomalies based on the estimation of uncertainty in neural networks. 
As one of the earlier works, Kendall~et~al.'s \textit{Bayesian SegNet}~\cite{kendallBayesianSegNetModel2016} derives the uncertainty of the \textit{semantic segmentation} (SemSeg) network \textit{SegNet} by Monte Carlo dropout sampling, where higher variance of the classes indicates higher uncertainty. The uncertainty can be interpreted as a pixel-wise anomaly score to detect obstacles on roads \cite{vojirRoadAnomalyDetection2021a, ohgushiRoadObstacleDetection2021}. A similar approach to detect unknown obstacles on the road is proposed by Jung~et~al.~\cite{jungStandardizedMaxLogits2021a}. They obtain class-conditioned standardized max logits of a segmentation network. This procedure is motivated by the finding that max logits have their own ranges for different predicted classes. The mean and standard deviations are thereby determined from the training samples. Thus, the standardization can be categorized as a \textit{learned confidence score} approach. In addition to the standardization, they suppress class boundaries and apply a dilated smoothing to consider local semantics in broad receptive fields. Heidecker~et~al.~\cite{heideckerCornerCaseDetection2021} model the epistemic uncertainty of \textit{Mask R-CNN} \cite{heMaskRCNN2017} and quantify the class and positional uncertainty of instances. They outline a criterion to detect anomalies based on the position and class uncertainty. Anomalies due to positional uncertainty are defined by the standard deviation of scaled bounding boxes exceeding a predefined threshold. In addition, instances are considered anomalous due to class uncertainty whenever the standard deviation of any class is above the predefined threshold. But Bayesian segmentation networks are slow in inference due to their multiple forward passes through the network with Monte Carlo dropout for each frame. Therefore, Huang~et~al.~\cite{huangEfficientUncertaintyEstimation2018} simulate the sampling procedure via \textit{region-based temporal aggregation} in frame sequences and retain the network's online capability. To ensure the correct uncertainty estimation of moving objects, the previous segmentation is warped via optical flow. Bevandić~et~al.~\cite{bevandicDenseOutlierDetection2021} present a multi-task network to simultaneously segment the input frame into semantics as well as output an anomaly probability map. The latter overrides the SemSeg whenever a probability exceeds a threshold to calibrate the confidence score when the model faces outliers. Most recently, Du~et~al.~\cite{duVOSLearningWhat2022} presented the general learning framework \textit{Virtual Outlier Synthesis} (VOS), which contrastively shapes the decision boundary of neural networks by synthesizing virtual outliers. At first, they estimate a class-conditioned multivariate Gaussian distribution in the penultimate latent space. Afterwards, outliers are sampled from a sufficiently small $\epsilon$-likelihood region of this learned distribution. These virtual outliers near the class-boundary encourage the model to form a compact decision boundary between \ac{ID} and \ac{OOD} data. Furthermore, they propose a novel training objective with free energy as an uncertainty measurement, where \ac{ID} data has negative and the virtual outliers positive energy. During inference, \ac{OOD} objects are detected with a \textit{logistic Regressor} based on the uncertainty score.

While the former approaches concentrate on anomalies on the object level, Breitenstein~et~al.~\cite{breitensteinDetectionCollectiveAnomalies2021} are the first to detect collective anomalies. They learn the normal quantity of class-instances based on a reference dataset. The class-instances themselves are predicted via a \textit{Mask R-CNN}~\cite{heMaskRCNN2017} to end up with a discrete distribution of classes. Furthermore, they introduce a variation of the \ac{EMD} for inference, namely the \ac{EMDEV}. Besides the comparison of distributions, the \ac{EMDEV} is a signed value which indicates whether the scene contains more or less instances of a class than usual.

\textbf{Reconstructive.} Reconstructive and generative approaches are predominantly used for anomaly detection on the object level, since the models learn to reproduce the normality of the training data without any auxiliary data of anomalous objects. For instance, a recent work by Vojir~et~al.~\cite{vojirRoadAnomalyDetection2021a} proposes the reconstruction module \textit{JSR-Net} to detect road anomalies based on a pixel-wise score. They enhance trained SemSeg networks by incorporating their information from known classes into the anomaly score. The network architecture consists of a reconstruction and a semantic coupling module. The former is connected to the backbone of the SemSeg network and reconstructs the road in a discriminative way, meaning it reduces the reconstruction loss of the road while increasing the loss for the remaining environment. In the subsequent module, the resulting pixel-error map is coupled with the output logits of the SemSeg to end up with a pixel-wise anomaly score. The extension module is trained on augmented road images, where patches of noise or a part of the input image are randomly positioned on the road and labeled as anomalous. The evaluation on various datasets shows the superiority of \textit{JSR-Net} in comparison to others \cite{lisDetectingUnexpectedImage2019,lisDetectingRoadObstacles2021, bevandicSimultaneousSemanticSegmentation2019, creusotRealtimeSmallObstacle2015} while preserving the closed-set segmentation performance. 

A similar approach is evaluated by Ohgushi~et~al.~\cite{ohgushiRoadObstacleDetection2021} against the \ac{LaF} benchmark on a highway dataset with real and synthetic road obstacles. In contrast to Vojir~et~al., they combine the entropy loss of the SemSeg with the perceptual loss between the real and reconstructed image to form an anomaly map. They outline a set of post-processing steps where the final obstacle score map depends on the semantic information, the aforementioned anomaly map, and a superpixel division to refine local regions. 

Di Biase~et~al.~\cite{dibiasePixelwiseAnomalyDetection2021a} leverage image re-synthesis~\cite{lisDetectingUnexpectedImage2019} by combining the reconstruction error with two uncertainty maps of the segmentation network. The network outputs the softmax entropy and distance additionally to the segmentation output. Similar to~\cite{ohgushiRoadObstacleDetection2021}, the perceptual difference is used as the reconstruction loss between the input and synthesized image. All predicted maps and the input image are fused in a spatial-aware dissimilarity module with three parts: encoder, fusion module, and decoder. In the fusion module, the encoded and re-synthesized inputs and the semantic image are concatenated and fused with a 1x1 convolution. The resulting feature map is evaluated against the jointly encoded uncertainty and perceptual difference via point-wise correlation. The final pixel-wise anomaly segmentation is provided by decoding the fused features and spatial-aware normalization with the semantic information. 

\textbf{Generative.} According to the \ac{FS}, \ac{LaF}, and \ac{SMIYC} obstacle track benchmarks, the dense anomaly detection with \textit{NFlowJS} of Grcić~et~al.~\cite{grcicDenseAnomalyDetection2021} outperforms all contemporary techniques and represents the current state-of-the-art of camera-based anomaly detection. \textit{NFlowJS} is jointly trained to generate synthetic negative patches with \textit{normalizing flows} (NF) atop regular images alongside training the dense prediction network based on these created mixed-content images. The generated negative patches are thereby defined as the anomaly mask. During training, the discriminative model is encouraged to yield a uniform predictive distribution for the generated patch. This induces the generative distribution of the NF to move away from the inliers. At the same time, it is trained to maximize the likelihood of inliers. These opposing objectives support the generation of images at the boundary of the training data while sensitizing the discriminative model for anomalies. Especially, due to the former facet, the synthesized anomaly patches are likely to contain parts similar to inliers where the model predicts with high confidence. A strong penalizing of this behavior demolishes the model's confidence on actual inlier pixels. Therefore, they find the \ac{JS} divergence as a mildly penalizing loss of high confidence predictions. During inference, the closed-set segmentation is masked by the anomaly map generated by a threshold exceeding temperature scaled softmax and the \ac{JS} divergence between output probability and uniform distribution. In contrast to former generative models, the \textit{NFlowJS} does only rely on the anomaly synthesis during training, resulting in a real-time inference speed. Blum~et~al.~\cite{blumFishyscapesBenchmarkMeasuring2021} also evaluate an NF based approach with logistic regression on their \ac{FS} benchmark. However, the results are incomparable with \textit{NFlowJS}. %As NFs are a bijection between the latent representation and generated image

Nitsch~et~al.~\cite{nitschOutofDistributionDetectionAutomotive2021} adopt and enhance a generative approach of Lee~et~al.~\cite{leeSimpleUnifiedFramework2018} for the detection of object anomalies. Lee~et~al.~propose an auxiliary \textit{generative adversarial network} (GAN) which encourages an object classifier to provide low confidence for samples outside the training distribution. Nitsch~et~al. extend the approach by a post hoc network statistic, which estimates a class-conditioned Gaussian distribution over the network's weights of the bottleneck layer. A cosine similarity metric determines the distribution distance and classifies a given sample based on an empirical threshold. Since they only perform classification, the localization of objects has to be done in advance.

Similarly, Lis~et~al.~\cite{lisDetectingUnexpectedImage2019} adopt GANs to re-synthesize the input image and detect anomalies on the object level by the difference in appearance. However, the image generation is based on the final SemSeg map in contrast to~ \cite{vojirRoadAnomalyDetection2021a, ohgushiRoadObstacleDetection2021}, where a decoder reconstructs the image based on the intermediate feature space of the SemSeg. As the SemSeg preserves the scene layout but loses the precise scene's appearance, regular reconstruction errors, like the perceptual loss, would output a high overall difference without informative results. Thus, they propose a discrepancy network which encodes the input and the re-synthesized image via multiple \textit{VGG16} \cite{simonyanVeryDeepConvolutional2015} networks with shared weights. The features are collectively concatenated with the convolutional encoded semantic map and correlated on all extraction levels and fed into the final decoding CNN on the respective feature level. The semantic-to-image synthesis is also adopted and evaluated by \cite{xiaSynthesizeThenCompare2020, haldimannThisNotWhat2019} in form of a \textit{conditional GAN} (cGAN) with a subsequent dissimilarity scoring.

%\textbf{Domain shift.}
Löhdefink~et~al.~\cite{lohdefinkSelfSupervisedDomainMismatch2020} present an approach for the detection of domain shifts. An autoencoder learns the domain of a given dataset in a self-supervised manner. The approach characterizes the training data domain via the distribution of the autoencoder's peak \ac{PSNR}. During inference, the \ac{DM} is estimated by comparing the learned and incoming \ac{PSNR} distribution of the data via the \ac{EMD}. The evaluation shows a strong rank order correlation between the autoencoder's \ac{DM} metric and the decrease of SemSeg performance when faced with target domains different than the source domain. While the inference is real-time capable, the approach has to  accumulate a certain number of images, as it uses batches as input.

\textbf{Feature Extraction.} 
Another domain shift detection is proposed by Bolte~et~al.~\cite{bolteUnsupervisedDomainAdaptation2019}, where the \ac{MSE} of feature maps is compared. The \ac{MSE} is evaluated over entire datasets or batches. Similarly, Zhang~et~al.~\cite{zhangDeepRoadGANbasedMetamorphic2018} propose the \textit{DeepRoad} framework to validate single input images based on the distance to the training embedding of \textit{VGGNet} features \cite{simonyanVeryDeepConvolutional2015}.
% Xue~et~al.~\cite{xueNovelMultilayerFramework2019} present a feature extraction approach which generates visual occlusion maps over multiple distances to detect even tiny road obstacles. For this purpose they train an obstacle-aware regressor, i.e., a \textit{random forest} (RF), to generate an obstacle occupied probability map under consideration of pseudo distances and edge features. 
Bai~et~al.~\cite{baiRecognizingAnomaliesUrban2018} detect anomalies in urban road scenes and classify entire input scenes as anomalous. They identify a set of representatives for normal urban scenes via the \textit{k-means clustering} of \textit{scale-invariant feature transform} (SIFT) features. Finally, images are classified by a \textit{one-class support vector machine} (one-class SVM).% based on the dense matching with the predefined scene references. %Due to the feature extraction and SVM, only a small number of training samples is required. 

% %Summary
Overall, many of the previously outlined techniques work without external data, but require a retraining of the proposed extension module or entire detection architecture.

%% file: sections/2_lidar.tex
\section{Anomaly Detection on Lidar Data}

% - Punktweise Outlier sind out-of-scope (kein denoising etc)
% - Fokus auf Objekt/Cluster-bezogene Outlier

% - lidar unter verschiednene Wetterbedingungen? JA (max 3 und ähnliche unter einem zusammenfassen und nur auf weitere verweisen) Goelles~et~al.~- 2020 - Fault Detection, Isolation, Identification and Rec.pdf
% - Ähnlich wie bei Camera eingrenzen (20 Wetter paper)
% - keine system diagnose
% - Wenige auf object eben sondern viel denoising

% - What about inference {An Experiment of Mutual Interference between Automotive lidar Scanners}
% - Introduction of openset-setting (known and unknown classes)
% - Motivation: Sensor degradation due to weather conditions like rain
% - No automotive column as lidar (as well as radar) is often used in automotive area

\begin{table*}[ht]
\caption{Overview of anomaly detection techniques on automotive lidar data}
\label{tab:lidarOverview}
\resizebox{\textwidth}{!}{
\begin{tabular}{@{}lcllllll@{}}
\toprule
Author(s) & Year & Ref & Technique  &  Approach  & Corner Case Level & Dataset / Simulation & Online\\ \midrule
Zhang~et~al.~&2021& \cite{zhanglidarDegradationQuantification2021} & DeepSAD \cite{ruffDeepSemiSupervisedAnomaly2020}& Confidence --- Learned & Domain  --- Domain Shift & Simulation \& static / dynamic real envs.& \xmark*\\
Cen~et~al.~&2021&\cite{cenOpenset3DObject2021} &MLUC&Confidence --- Learned&Object  --- Single-Point &UDI\cite{cenOpenset3DObject2021} \& KITTI\cite{geigerVisionMeetsRobotics2013}&\xmark*\\
Wong~et~al.~&2019&\cite{wongIdentifyingUnknownInstances2019} &OSIS&Confidence --- Learned&Object  --- Single-Point &TOR4D\cite{wongIdentifyingUnknownInstances2019} \& Rare4D&\xmark*\\
Masuda~et~al.~& 2021&\cite{masudaUnsupervised3dPoint2021} &VAE (FoldingNet) &Reconstruction&Object  --- Single-Point &ShapeNet\cite{changShapeNetInformationRich3D2015}&\xmark*\\
\bottomrule
\end{tabular}}
\end{table*}
Most often, autonomous vehicles do not solely rely on camera data. Although, camera data has the highest resolution of the three sensor modalities, it lacks an accurate measurement of depth. Therefore, \textit{light detection and ranging} (lidar) sensors, which provide a three-dimensional depth map of the environment, are often found in sensor setups. While there is much research about local denoising of lidar point clouds on the pixel level \cite{regayaPointDenoiseUnsupervisedOutlier2021, baltaFastStatisticalOutlier2018}, we are interested in anomalies on object and domain level, where an entire cluster of points or a large and constant shift in appearance is considered as anomalous. Especially weather conditions like rain, snow, and fog heavily influence the data. All lidar-based methods can be found in Table~\ref{tab:lidarOverview}.

\textbf{Confidence score.} Recent research by Zhang~et~al.~\cite{zhanglidarDegradationQuantification2021} shows that rain affects the lidar measurement quality, as resulting point clouds are sparser, noisier, and the average intensity is lower. Therefore, they aim to quantify the lidar degradation with the \textit{Deep Semi-supervise Anomaly Detection} (DeepSAD) approach~\cite{ruffDeepSemiSupervisedAnomaly2020}. They first project 3D lidar data into a 2D intensity image. DeepSAD then transforms the images into a latent space, where all normal images, i.e., the scans without rain, fall into a hypersphere and all abnormal, i.e., rain affected, images are mapped away from the hypersphere's center. Finally, the distance of a transformed test image to the learned center of the hypersphere is interpreted as the anomaly score. As the model architecture defines anomalies as those who fall out of the hypersphere, we list the proposed methodology as a \textit{learned confidence} detection approach in Table~\ref{tab:lidarOverview}. The trained DeepSAD reaches a Spearman’s correlation of up to 0.82 between the rainfall intensity and degradation score on dynamic, simulated test data. This indicates a considerably accurate quantification of anomaly detection due to weather conditions. Although the approach is developed for rainy and normal weather conditions, we suspect that the proposed method is transferable to other weather conditions, such as snow and fog. 
% Furthermore, we consider the method necessary to develop vehicles with level 3+ autonomy which implies that the responsibility for the perception of the environment is transferred to the vehicle \cite{goellesFaultDetectionIsolation2020}. Hence, autonomous vehicles have to deal with scenes under extreme weather conditions as humans do, by e.g. reducing the speed or increasing the following headway. The ability to initiate such precautionary actions might be imitated by constantly monitoring metrics like the aforementioned degradation score. Also weather conditions which are abnormal or at least unusual for specific regions, like sandstorms, might trigger a higher degradation score indicating to change to a more conservative driving style. 

In the past, several architectures have been proposed to detect objects in point clouds, like \textit{VoxelNet}~\cite{zhouVoxelNetEndtoEndLearning2018}, \textit{PointRCNN}~\cite{shiPointRCNN3DObject2019}, and \textit{PointNet++}~\cite{qiPointNetDeepHierarchical2017}. However, these are based on a closed-set setting, thus being only capable of detecting classes that were included in the training set. In contrast, open-set detection methods are able to explicitly classify objects outside the closed-set as unknown upon the regular detection of the predefined classes. The open-set setting therefore loosens the constraint to classify all detections as one of the predefined classes. Consequently, one expects the false positive rate to improve and the model to acknowledge the novelty of objects upon never seen instances.

The idea of an open-set detector for 3D point clouds was first implemented by Wong~et~al.~\cite{wongIdentifyingUnknownInstances2019}. They propose an \textit{Open-Set Instance Segmentation} (OSIS) network, which learns a category-agnostic embedding to cluster points into instances regardless of their semantics. The inference is based on a \ac{BEV} lidar frame and consists of two stages: the closed-set and open-set perception. In the first stage, a backbone of 2D convolutions extracts multi-scale features, which are then fed into a detection and an embedding head. The latter is the core of \textit{OSIS} and learns the category-agnostic embedding space. Moreover, the embedding head yields the prototypes of possible closed-set classes. Points are then associated to prototypes of known categories by the learned embedding space. In the second stage, the remaining unassociated points are considered as unknown. Those are clustered into instances of unknown objects via \textit{density-based spatial clustering of applications with noise}~(DBSCAN)~\cite{esterDensityBasedAlgorithmDiscovering1996}. The outlined approach falls into the category of \textit{learned confidence} scores, as the prototypes are learned during training and unknown objects are identified by their uncertainty of class association.
\textit{OSIS} is evaluated on two large-scale, non-public datasets. Here, the technique outperforms other adapted deep learning based instance segmentation algorithms for the detection of single-point anomalies on the object level. 

The \textit{OSIS} network is later used as a baseline for comparison of the \textit{Metric learning with Unsupervised Clustering} (MLUC) network developed by Cen~et~al.~\cite{cenOpenset3DObject2021}. They focus on two primary challenges: identifying regions of unknown objects with high probability and enclosing these regions' points with proper bounding boxes. In context of the first problem, the paper shows that the \ac{EDS}, based on metric learning, is more suitable than a naive softmax probability metric to differentiate between regions of known and unknown objects. They replace the classifier of closed-set detections with the euclidean distance representation to all prototypes of the embedding space. The euclidean distance-based probability is incorporated into the loss function, such that the embedding vector of known classes is close to the corresponding prototypes of the respective class. However, unknown objects are mapped close to the center of the embedding, having a smaller \ac{EDS}. The \ac{EDS} measures the uncertainty of closed-set detections. Therefore, boxes with an \ac{EDS} lower than a threshold $\lambda_{EDS}$ are considered as regions of unknown objects. Similarly to \textit{OSIS}, these bounding boxes of low confidence are then refined by unsupervised depth clustering. The \textit{MLUC} considerably outperforms \textit{OSIS}. %Furthermore, the authors show that the \textit{OSIS} network is even outperformed by the naive softmax probability 3D object detector on both datasets. 

\textbf{Reconstructive.} Masuda~et~al.~\cite{masudaUnsupervised3dPoint2021} show an  approach to detect whether an object point cloud is anomalous or not. In contrast to the preceding methods, this technique is based on point clouds of single encapsulated objects. Since automotive lidars provide full environment scans, single objects or regions of interest would need to be extracted by detection or clustering approaches first. % This means, the architecture is neither able to localize abnormal objects nor to detect multiple abnormal objects in an entire environment scan. But on the contrary, the proposed \textit{variational autoencoder} (VAE) architecture is able to learn a wider range of objects as normal in its latent space in an unsupervised manner. The approach might be useful for a subsequent evaluation of detections with low confidence regarding an abnormal shape.
The proposed VAE is based on the \textit{FoldingNet} decoder \cite{yangFoldingNetPointCloud2018} and learns to reconstruct the set of known objects which are considered as normal. The point cloud is then classified as anomalous based on the reconstruction and the Chamfer distance as an anomaly score. The approach is evaluated on the ShapeNet~\cite{changShapeNetInformationRich3D2015} dataset, which also includes a variety of objects outside the \ac{AD} domain. The results are promising, as the model achieves an average AUC of 76.3\%, where known classes were defined as anomalies.

%Summary & findings of lidar anomaly detection
% In summary, as lidars are frequently used in the domain of \ac{AD}, many approaches are designed for automotive data. However, Table~\ref{tab:lidarOverview} shows that the development of anomaly detection techniques for lidars beyond the pixel level lags behind and is only slowly gaining momentum. 
Overall, anomaly detection on the object level in lidar data is just gaining momentum, after research has already led to various closed-set detection architectures.

%% file: sections/3_radar.tex
\section{Anomaly Detection on Radar Data}

% - Fokus auf Radare im Automotive Sektor (entweder verfügbar oder in Forschung), kein Maritime
% - no UWB Radars
% - No Noise reduction techniques
% - No through-the-wall radars 
% - Multi-path effect due to ground surface??? Nein
% - Sind ghost targets anomalien? Ja (sehr sensor spezifisch)
% \begin{itemize}
%     \item Ghost target modeling and simulation \cite{holderModelingSimulationRadar2019}
%     \item Categorize into mathmatical multipath modeling \& Simulation of ghost targets and DL ML based approaches
%     \item Connect to next section by refering to self attention in multi-modal approach
%     \item interesting at what range the dataset/simulation is evaluated as radars advantage is especially the long-distance and robust measurment 
%     \item outline future work (temporal information to be considered)
%     \item higher radar freqeuncey and more data (pblic ghost target labeled dataset) \cite{chamseddineGhostTargetDetection2021}
% \end{itemize}

\begin{table*}[ht]
\caption{Overview of anomaly detection techniques on automotive radar data}
\label{tab:radarOverview}
\resizebox{\textwidth}{!}{
\begin{tabular}{@{}lclllllll@{}}
\toprule
Author(s) &Year& Ref & Technique  &  Approach & Method set & Corner Case Level & Dataset / Simulation & Online \\ \midrule
Liu~et~al.~&2021& \cite{liuMultipathPropagationAnalysis2021} & Range difference & Feature Extraction &Mathematical& Scene --- Contextual  & Numerical simulation ($\leq50m$) & \cmark\\
Griebel~et~al.~&2021& \cite{griebelAnomalyDetectionRadar2021} & MFG PointNet++\cite{qiPointNetDeepHierarchical2017} & Feature Extraction &DL&\begin{tabular}{@{}l@{}}Scene --- Contextual \\ (Pixel --- Local Outlier)\end{tabular} &Hand-labeled 2D data ($\leq70m$) &\cmark (42.7 FPS)\\
Chamseddine~et~al.~&2021&\cite{chamseddineGhostTargetDetection2021} & PointNet \cite{qiPointNetDeepHierarchical2017}&Feature Extraction&DL&Scene --- Contextual&Lidar-labeled 3D data& \cmark**\\
Kraus~et~al.~&2020& \cite{krausUsingMachineLearning2020} & PointNet++ \cite{qiPointNetDeepHierarchical2017}& Feature extraction & DL & Scene --- Contextual & NLOS\cite{scheinerSeeingStreetCorners2020}& \cmark**\\
Prophet~et~al.~&2019& \cite{prophetInstantaneousGhostDetection2019}&Features + RF&Feature Extraction& Feat. Eng. \& ML &Scene --- Contextual&Hand-labeled 2D data &\cmark\\
%Holder~et~al.~& \cite{holderModelingSimulationRadar2019} & &&&& &\xmark\\
Ryu~et~al.~&2018& \cite{ryuDetectingGhostTargets2018} &MLP&Feature Extraction&ML&Scene --- Contextual&City center intersection&\cmark\\
Kamann~et~al.~& 2018&\cite{kamannAutomotiveRadarMultipath2018} & Geometric propagation&Feature Extraction&Mathematical&Scene --- Contextual&Experimental setup &\cmark\\
Visentin~et~al.~&2017& \cite{visentinAnalysisMultipathDOA2017} &Pauli decomposition&Feature Extraction&Mathematical&Scene --- Contextual & Experimental setup & \cmark\\
Roos~et~al.~&2017& \cite{roosGhostTargetIdentification2017} & Orientation \& motion&Feature Extraction&Mathematical&Scene --- Contextual&Simulation&\xmark\\
Garcia~et~al.~&2019& \cite{garciaIdentificationGhostMoving2019} & OGM + CNN&Reconstructive&DL&Scene --- Contextual&Hand-labeled 2D data&\xmark\\
\bottomrule
\end{tabular}}
\end{table*}
%Why radar (brief comparison to previous modalities)
Radar is the third sensor modality often used in \ac{AD}. It has a higher range at the cost of a lower resolution and less detailed spatial information than lidar sensors. In comparison to both previous modalities, radar is more robust to changing weather and daytime conditions \cite{wangRadarGhostTarget2021}. In the following, we concentrate on anomaly detection techniques designed for radar systems installed in the automotive industry, like surround, long, and short range radars and exclude techniques based on ultra-wideband and through-the-wall radars. We additionally characterize approaches by the method set (mathematical, feature engineering, \ac{ML} or \ac{DL}) used for detection. All radar-based methods can be found in Table~\ref{tab:radarOverview}.

%Ghost targets as one of the biggest challenge affecting realization of AD
Radar estimates an objects' position by measuring the time of flight of electromagnetic multipath waves and their reflections. Due to the multipath propagation, radar can detect even occluded objects~\cite{thaiAroundthecornerRadarDetection2017}. However, this advantage is mitigated by the fact that this also causes noise, reflections and artifacts. Especially reflective surfaces, like guardrails on highways or smooth walls, produce non-existing artifacts, often refereed to as “ghost targets”~\cite{wangRadarGhostTarget2021, ryuDetectingGhostTargets2018, chamseddineGhostTargetDetection2021}. These are a long-standing challenge affecting automotive radars~\cite{liuMultipathPropagationAnalysis2021}. For this reason, and as this survey focuses on anomalies above the pixel level, we specifically concentrate on methods to detect ghost targets and alike. 
% %Ghost targets Typisierung zur Einordnung
% Ghost targets can be classified into two types of multi-path reflections following the convention of Jiangang~et~al.~\cite{liuFirstOrderMultipathGhosts2016}. Those where the last bounce happens on the real object and those where it bounces off the reflective surface are referred to as type-1 and type-2 reflections, respectively. Moreover, the nomenclature categorizes reflections by the number of objects they bounce off as first-, second- or higher-order bounce detections. Most of the reviewed literature \cite{krausUsingMachineLearning2020, prophetInstantaneousGhostDetection2019, garciaIdentificationGhostMoving2019} concentrate on type-2 ghost targets and rather low-order bounces as the received signal energy decreases with the number of bounces resulting in a rapid decrease of ghost detections.

%Work based on mathematical modeling and simulating ghost targets as baseline approaches
\textbf{Feature Extraction.} Most recent work by Liu~et~al.~\cite{liuMultipathPropagationAnalysis2021} proposes a model of multipath propagation to identify and remove ghosts based on the targets' range difference, based on reflections from a guardrail. The established model and numerical results show that the range difference between each real vehicle and its corresponding ghost target only differs slightly. In contrast, the range differences between two, even closely located, real targets are usually far greater. The proposed ghost removal algorithm leverages this finding as it distinguishes between real and ghost targets based on a maximum range difference threshold $\Delta{r}$, which is numerically determined in advance.     
While this mathematical approach is simple and effective in simulation, one has to consider its constraints, as it is limited to a highway-like driving scene with three lanes of fixed size. Moreover, the distance between the target and the reflective guardrail takes only three values and does not simulate lane changes of real vehicles. Similar work was done by Holder~et~al.~\cite{holderModelingSimulationRadar2019}, Kamann~et~al.~\cite{kamannAutomotiveRadarMultipath2018}, Visentin~et~al.~\cite{visentinAnalysisMultipathDOA2017}, and Roos~et~al.~\cite{roosGhostTargetIdentification2017}.

%ML/DL based approaches
The latest \ac{ML} algorithms are utilized to detect radar anomalies in a greater variety of driving scenes without the aforementioned constraints of a mathematical model to work. In this context, ghost targets are often defined as a separate class. For instance, Griebel~et~al.~\cite{griebelAnomalyDetectionRadar2021} implement a \ac{DL} method utilizing the \textit{PointNet++} architecture. The original architecture uses \ac{MSG} layers to extract features on different scales in a point cloud. The \ac{MSG} module uses a circular form to query a point's neighboring information. They introduce an extension of the original grouping module, hypothesising that anomalous radar targets occur in a ring-shaped region around the radar sensor origin within the same range as car targets. The so-called multi-form grouping (MFG) module is a combination of the original circular as well as the new ring querying form. Hence, the module incorporates the neighborhood information of both forms at multiple scales. Moreover, they do not solely focus on the detection of multi-path anomalies, like ghost targets, but also on other single target anomalies caused by the Doppler velocity ambiguities or errors in the direction of arrival estimation. The latter are local outliers and fall into the pixel level. 

Kraus~et~al.~\cite{krausUsingMachineLearning2020} utilize \textit{PointNet++} to not only differentiate between real and ghost objects, but also classify them as (ghost) pedestrians or (ghost) cyclists. Therefore, the evaluation is limited to the \ac{NLOS} dataset~\cite{scheinerSeeingStreetCorners2020}, including only vulnerable road users. They tackle the challenge of sparse radar data by accumulating measurements over a period of 200ms.

While the former approaches detect anomalies in single-shot 2D radar data, Chamseddine~et~al.~\cite{chamseddineGhostTargetDetection2021} evaluate the \textit{PointNet++} architecture to detect ghost targets in dense 3D radar data. The \textit{PointNet++} architecture is, in contrast to other common 3D detection networks~\cite{zhouVoxelNetEndtoEndLearning2018,langPointPillarsFastEncoders2019}, able to learn individual point features and therefore well suited to classify single radar points into real or ghost targets. Ablation studies show that the form of representation of spatial information matter as the additional encoding of points in spherical coordinates boosts the network's performance. % The spherical coordinate system allows capturing the connections between points more easily, as e.g. points with the same azimuth and elevation only differ in depth, unlike in cartesian coordinates where the distance would be the square root of the sum of squares. 
% Additionally, this representation seems to be more natural as the sensor collects data in spherical coordinates.

\begin{table*}[ht]
\caption{Overview of anomaly detection on multimodal sensor data}
\label{tab:multi_overview}
\resizebox{\textwidth}{!}{
\begin{tabular}{@{}lclllllll@{}}
\toprule
Author(s) & Year &Ref & Technique  & Approach & Corner Case Level& Dataset / Simulation & Online\\ \midrule
Wang~et~al.~&2021& \cite{wangRadarGhostTarget2021} & Multimodal transformers & Feature Extraction & Scene --- Contextual & Auto-labeled nuScenes\cite{caesarNuScenesMultimodalDataset2020} & \xmark* \\
Sun~et~al.~& 2020&\cite{sunRealtimeFusionNetwork2020}& RGB-D network&Feature Extraction&Scene --- Contextual&\ac{CS}\cite{cordtsCityscapesDatasetSemantic2016}&\cmark (22 FPS)\\
Ji~et~al.~&2020& \cite{jiMultiModalAnomalyDetection2020} & SVAE & Feature Extraction & Scene --- Contextual & TerraSentia & \cmark \\
% Ravanbakhsh~et~al.~& \cite{ravanbakhshLearningMultiModalSelfAwareness2018} &&&&&&\\
Gupta~et~al.~&2018& \cite{guptaMergeNetDeepNet2018} & MergeNet & Feature Extraction & Scene --- Contextual & LaF\cite{pinggeraLostFoundDetecting2016a} & \xmark (5 FPS)\\
Pinggera~et~al.~&2016& \cite{pinggeraLostFoundDetecting2016} & FPHT & Feature Extraction & Scene --- Contextual & LaF\cite{pinggeraLostFoundDetecting2016a} & \cmark (20 FPS)\\
\bottomrule
\end{tabular}}
\end{table*}

Another noteworthy approach to detect ghost anomalies regardless of their causes is the procedure of Prophet~et~al.~\cite{prophetInstantaneousGhostDetection2019}, where initially moving targets are identified by the scanned radial velocity and a threshold value to improve scene understanding. Afterwards, a set of handcrafted features is defined for each \ac{DUT}. These features comprehend the \ac{DUT} parameters, the vehicles motion state, the error value calculated in the first step, the number of static and moving neighbors, as well as the calculation of an \ac{OGM} around the \ac{DUT}. Moreover, they include a Boolean feature indicating the presence of a moving neighbor detection around the \ac{DUT} in the previous frame. Consequently, this technique is the first to incorporate temporal data to improve detection. Finally, these features are fed into \ac{ML} algorithms like a SVM, \textit{k nearest neighbor classifier} (KNN), or RF. According to the subsequent evaluation on a data set of 36,916 detections, the RF outperforms all other algorithms with a success rate of 91.2\%. Similarly, Ryu~et~al.~\cite{ryuDetectingGhostTargets2018} train a \textit{multilayer perceptron} (MLP) on a set of six features to remove ghost targets from a tracking algorithm. 

\textbf{Reconstructive.} Garcia~et~al.~\cite{garciaIdentificationGhostMoving2019} use a two-channel image consisting of the aforementioned occupancy grid and moving detections map as an input of a \textit{fully convolutional network} (FCN). The proposed architecture is segmented in an encoder and a decoder part. While the former extracts the semantic information into a lower resolution representation, the latter reconstructs the spatial information and maps the extracted representation back to the original image size. In the resulting map of probabilities, a  moving target is considered a ghost detection. The technique achieves a binary classification accuracy of 92\% on a test set of 50 images.

% Overall, current techniques for the detection of anomalies in radar sensors are difficult to compare, as the community misses a common benchmark with labeled ghost targets. However, promising labeling tools have been developed based on available lidar data, which is less susceptible to reflections~\cite{chamseddineGhostTargetDetection2021}. 
Overall, many approaches assume that ghost and real targets can be differentiated by their feature set in contrast to conventional, i.e., reconstructive or confidence based techniques as shown in Table~\ref{tab:radarOverview}. Despite that, we expect future work to further improve by taking into account temporal information, as indicated in \cite{prophetInstantaneousGhostDetection2019}.

%% file: sections/4_multimodal.tex
\section{Anomaly Detection on Multimodal Data}

% \begin{table*}[ht]
% \caption{Overview of anomaly detection on multimodal sensor data}
% \label{tab:multi_overview}
% \resizebox{\textwidth}{!}{
% \begin{tabular}{@{}lclllllll@{}}
% \toprule
% Author(s) & Year &Ref & Technique  & Approach & Corner case & Dataset/Simulation & Online\\ \midrule
% Wang~et~al.~&2021& \cite{wangRadarGhostTarget2021} & Multimodal transformers & Feature extraction & Scene level --- Contextual & Auto-labeled nuScenes\cite{caesarNuScenesMultimodalDataset2020} & \xmark* \\
% Sun~et~al.~& 2020&\cite{sunRealtimeFusionNetwork2020}& RGB-D network&Feature extraction&Scene level --- Contextual&\ac{CS}\cite{cordtsCityscapesDatasetSemantic2016}&\cmark (22 FPS)\\
% Ji~et~al.~&2020& \cite{jiMultiModalAnomalyDetection2020} & SVAE & Feature extraction & Scene level --- Contextual & TerraSentia & \cmark \\
% % Ravanbakhsh~et~al.~& \cite{ravanbakhshLearningMultiModalSelfAwareness2018} &&&&&&\\
% Gupta~et~al.~&2018& \cite{guptaMergeNetDeepNet2018} & MergeNet & Feature extraction & Scene level --- Contextual & LaF\cite{pinggeraLostFoundDetecting2016a} & \xmark (5 FPS)\\
% Pinggera~et~al.~&2016& \cite{pinggeraLostFoundDetecting2016} & FPHT & Feature extraction & Scene level --- Contextual & LaF\cite{pinggeraLostFoundDetecting2016a} & \cmark (20 FPS)\\
% \bottomrule
% \end{tabular}}
% \end{table*}

Autonomous vehicles are typically equipped with multiple modalities. In the following, we provide an overview of techniques which identify anomalies based on irregularities between the individual sensors or by fusing information. All multimodal methods can be found in Table~\ref{tab:multi_overview}.

\textbf{Feature Extraction.} Following on from the previous detection of ghost targets in radar data, Wang~et~al.~\cite{wangRadarGhostTarget2021} propose a multimodal technique. Transformers are well suited for 3D point clouds as their attention mechanism is permutation invariant, which is hard for conventional neural networks. Moreover, transformers explicitly model a point's interactions, in contrast to the aforementioned architectures like \textit{PointNet++}. The authors adopt a multimodal transformer network to detect radar ghost targets by referencing lidar points. Radar point clouds are way sparser than lidar point clouds, which hinders the data matching. Therefore, individual radar points query for surrounding lidar points by KNN and provide local feature information, like a “magnifying lens”. They apply self-attention for the unstructured radar data itself to identify ghost targets, as these show high affinity to the corresponding real targets. The attention modules are stacked to a network. Lastly, the fully connected segmentation head of \textit{PointNet++} is utilized to classify individual radar points as possible ghost targets. The proposed method is evaluated on the nuScenes datatset~\cite{caesarNuScenesMultimodalDataset2020}. Worth mentioning, the ground truth of ghost targets was generated by comparing radar and lidar data.

Sun~et~al.~\cite{sunRealtimeFusionNetwork2020} present a real-time fusion network for SemSeg based on RGB-D data. The primary goal of the multimodal architecture is to improve image segmentation by incorporating depth information. Furthermore, they argue that the multi-source segmentation framework is also capable to detect unexpected road obstacles, providing a unified pixel-wise scene understanding. However, the evaluation on the \ac{CS} dataset \cite{cordtsCityscapesDatasetSemantic2016} does not provide detection performance measures for the unexpected obstacles, as the approach concentrates on the SemSeg of closed-set classes. Another RGB-D based detection of road obstacles is implemented by Gupta~et~al.~\cite{guptaMergeNetDeepNet2018} in form of \textit{MergeNet}. As the architecture's name suggests, the model merges two networks, the \textit{Stripe-net} and \textit{Context-net}, via a third meta \textit{Refiner-net}. The \textit{Stripe-net} extracts low-level features of the RGB and depth data in parallel, based on images split in stripes. This forces the network to learn discriminative features within narrow bands of information and a small subset of parameters. Moreover, this allows for a more reliable detection of small road obstacles. In contrast, the \textit{Context-net} is trained on the entire RGB image and is determined to learn high-level features. The \textit{Refiner-net} acts as a meta network to combine the complementary features and end up with a form of curriculum learning. As a result, \textit{MergeNet} is trained to discriminate between road, off-road, and small obstacle, where we consider the latter as abnormal. 

Ji~et~al.~\cite{jiMultiModalAnomalyDetection2020} propose a \textit{supervised VAE} (SVAE) to merge multiple sensor modalities of different dimensionality. This is especially useful for the fusion of dense lidar data and radar data of lower resolution. They abandon the decoder after training and use the learned encoder as a feature extractor. The modalities' latent representation is then – along with other encoded modalities – fed into a fully connected layer to identify an anomalous operation mode of the vehicle. Even though the method was designed for field robots, we expect it to be transferable to other driving scenes.

% A well known road obstacle dataset is the \textit{Lost and Found} dataset published by Pinggera~et~al.~\cite{pinggeraLostFoundDetecting2016}. Together with the publication of the dataset, the authors also propose and evaluate a method for the detection of road obstacles based on stereo camera images. The authors extend the point-wise obstacle detection of Pinggera~et~al.~\cite{pinggeraHighperformanceLongRange2015} by a mid-level representation in form of Cluster-Stixels (CStix) to increase performance robustness and reduce computational cost. 
In summary, one can see in Figure~\ref{fig:bar_chart_approaches}, that all of the multimodal anomaly detection techniques are based on the comparison of the individual modalities' \textit{extracted features}. We argue, that multimodal detection could become much more relevant, as it broadens the search space for potential anomalies, while reducing the risk of false positives. 

%% file: sections/5_object.tex
\section{Anomaly Detection on Abstract Object Data}

\begin{table*}[ht]
\caption{Overview of anomaly detection techniques on abstract object level data}
\label{tab:abstraction_object_overview}
\resizebox{\textwidth}{!}{
\begin{tabular}{@{}lcllllll@{}}
\toprule
Author(s) &Year& Ref & Technique & Approach  & Corner Case Level& Dataset / Simulation& Online\\ \midrule
Yang~et~al.~&2019&\cite{yangDrivingBehaviorAssessment2019}&HMM&Prediction&Scenario --- Risky &CARLA\cite{dosovitskiyCARLAOpenUrban2017}& \cmark\\
Bolte~et~al.~&2019&\cite{bolteCornerCaseDetection2019}&Adversarial AE&Prediction&Scenario --- Anomalous, Novel, Risky&\ac{CS}\cite{cordtsCityscapesDatasetSemantic2016}&\xmark*\\
Liu~et~al.~&2018&\cite{liuFutureFramePrediction2018}&U-Net \cite{ronnebergerUnetConvolutionalNetworks2015} + Flownet \cite{dosovitskiyFlowNetLearningOptical2015} &Prediction &Scenario --- Anomalous, Novel & CUHK\cite{luAbnormalEventDetection2013a}, UCSD\cite{mahadevanAnomalyDetectionCrowded2010}, ST\cite{luoRevisitSparseCoding2017}&\cmark (25 FPS)\\
%Zhang~et~al.~&\cite{zhangSafeDriveOnlineDriving2017} & SafeDrive &&&\\
Yuan~et~al.~&2018&\cite{zhangDeepRoadGANbasedMetamorphic2018} &Bayes Model&\begin{tabular}{@{}l@{}}Prediction\\Confidence --- Bayesian\end{tabular}&  Scenario --- Anomalous&Driving videos& \xmark\\
Zhang~et~al.~&2018&\cite{zhangDeepRoadGANbasedMetamorphic2018} & DeepRoad & Feature Extraction& Scenario --- Anomalous& Udacity\cite{FinalLeaderboardUdacity2016} & \cmark\\
Stocco~et~al.~& 2020&\cite{stoccoMisbehaviourPredictionAutonomous2020} & SelfOracle & \begin{tabular}{@{}l@{}}Reconstructive\\Confidence --- Learned\end{tabular}& Scenario --- Risky & Udacity\cite{FinalLeaderboardUdacity2016} & \cmark \\
\bottomrule
\end{tabular}}
\end{table*}
The previous sections gave an overview of anomaly detection techniques suitable for specific sensor modalities. The following approaches are focusing on a more abstract level of pattern analysis, i.e., the detection of anomalous behavior in scenarios, which are not necessarily bound to a sensor modality. Thus, the approaches are designed to detect anomalies on the scenario level \cite{breitensteinSystematizationCornerCases2020} and deal with risky and abnormal driving behavior of non-ego vehicles. All abstract object-level based methods can be found in Table~\ref{tab:abstraction_object_overview}.

\textbf{Prediction.} Yang~et~al.~\cite{yangDrivingBehaviorAssessment2019} assess the behavior of driving vehicles based on \textit{Hidden Markov Models} (HMM) to detect anomalous scenarios. The observation states of the Markov model are provided by the \textit{Conditional Monte Carlo Dense Occupancy Tracker} (CMCDOT) framework~\cite{rummelhardConditionalMonteCarlo2015} and comprise real-time velocity as well as vehicle position through probabilistic occupancy grids. The framework derives these observations based on point cloud and odometry data. As a result, the pipeline can reliably infer risky and abnormal driving behaviors in simulated multi-lane highway scenarios with two non-ego vehicles. % Therefore, the performance of the assessment pipeline in other scenes remains open.

Bolte~et~al.~\cite{bolteCornerCaseDetection2019} propose an anomaly detection on the scenario level, where patterns are observed over a sequence of sensor data, i.e., camera images. They consider all subtypes: anomalous, novel, and risky scenarios~\cite{breitensteinSystematizationCornerCases2020}. They quantify the anomalous behavior for moving objects, such as pedestrians or cars, due to the nature of scenario anomalies. The error between the real and a predicted frame is considered as the anomaly score. The predicted frame is generated by an adversarial autoencoder and based on the past sequence of input frames. Hence, the anomaly score can also be interpreted as the non-predictability of the model. The model is evaluated with \ac{MSE}, \ac{PSNR}, and \ac{SSIM}~\cite{wangImageQualityAssessment2004} metrics, and anomalous scenarios are determined by a threshold. They localize anomalous behaving objects by dividing the input image into grid cells of user-specific size and weight close objects higher, as those pose a higher risk of collision.

A similar, but more comprehensive, approach is outlined in the paper of Liu~et~al.~\cite{liuFutureFramePrediction2018}. They adopt \textit{U-Net}~\cite{ronnebergerUnetConvolutionalNetworks2015} as an image-to-image translation model to predict the next frame based on the past sequence of frames. In contrast to the former approach \cite{bolteCornerCaseDetection2019}, their framework considers also temporal information of scenarios. They extend their objective function by an optical flow constraint to retain the motion information of moving objects. The optical flow is calculated via \textit{Flownet}~\cite{dosovitskiyFlowNetLearningOptical2015}.
They leverage adversarial training to discriminate between real and fake images to further boost the performance of the future frame prediction. Anomalous scenarios are again identified by the \ac{PSNR} of the real and predicted frame exceeding a predefined threshold.

\textbf{Reconstructive.} Stocco~et~al. propose \textit{SelfOracle}~\cite{stoccoMisbehaviourPredictionAutonomous2020} for the detection of safety-critical misbehavior, like collisions and out-of-bound episodes. The architecture uses a VAE to reconstruct a set of preceding input images of a current scene and calculates the corresponding reconstruction errors. During the training on normal data, the model is fitting a probability distribution to the observed reconstruction errors via maximum likelihood estimation. The estimated distribution can then be used to determine a threshold value $\theta$ to distinguish between anomalous and normal behavior. The parameter $\epsilon$ corresponds to the probability of the tail and thus $\theta$ controls for the false positive rate of the detection. In addition, \textit{SelfOracle} implements a time-aware anomaly scoring by applying a simple autoregressive filter on the sequence of reconstruction errors, as the current error might be susceptible to single-frame outliers. While they evaluate \textit{SelfOracle} only in a simulation environment, the approach seems promising and even outperforms the author's implementation of the \textit{DeepRoad} framework. 

Finally, anomaly detection on the object level heavily depends on human driving behavior. Therefore, with the rise of autonomous vehicles on the road, \ac{AD} will experience a large concept drift in behavior prediction.

%% file: sections/6_conclusion.tex
\section{Conclusion}
% Gesamtüberblick über alle Modalitäten hinweg
In this paper, we provide an extensive survey of techniques for the detection of anomalies in the field of autonomous driving. While the survey by Breitenstein et al. that we build upon~\cite{breitensteinCornerCasesVisual2021} is limited to camera data, we characterize techniques across different sensor modalities. Most of the recent advancements are concerned with image-based anomaly detection, while lidar- and radar-based approaches are still struggling to gain momentum. One reason for this is the absence of benchmarks, which so far only exist in the camera sector. The community misses common datasets of labeled anomalies, which leaves the unified comparison of detection techniques difficult. Tables~\ref{tab:cameraOverview}-\ref{tab:abstraction_object_overview} show that each modality might be more suitable for the detection of one or only few types of corner cases, as e.g., lidar-based techniques focus strongly on single-point anomalies. Overall, the state-of-the-art especially detects contextual anomalies on the scene level, while collective anomalies lack behind. 